%% file: aaai25.tex
\documentclass[letterpaper]{article} 
\usepackage{aaai25}  
\usepackage{times}  
\usepackage{helvet}  
\usepackage{courier}  
\usepackage[hyphens]{url}  
\usepackage{graphicx} 
\usepackage{natbib}  
\usepackage{caption} 
\usepackage{booktabs}
\usepackage{algorithm}
\usepackage{algorithmic}
\usepackage{amsmath}
\usepackage{color}
\usepackage{multirow}

\usepackage{newfloat}
\usepackage{listings}
\DeclareCaptionStyle{ruled}{labelfont=normalfont,labelsep=colon,strut=off} 
\floatstyle{ruled}
\newfloat{listing}{tb}{lst}{}
\floatname{listing}{Listing}
\title{Unicorn: Unified Neural Image Compression with One Number Reconstruction}
\author{
    Qi Zheng\textsuperscript{\rm 1}\equalcontrib,
    Haozhi Wang\textsuperscript{\rm 1}\equalcontrib,
    Zihao Liu\textsuperscript{\rm 2},
    Jiaming Liu\textsuperscript{\rm 1},
    Peiye Liu\textsuperscript{\rm 2},
    Zhijian Hao\textsuperscript{\rm 1},\\
    Yanheng Lu\textsuperscript{\rm 2},
    Dimin Niu\textsuperscript{\rm 2},
    Jinjia Zhou\textsuperscript{\rm 3},
    Minge Jing\textsuperscript{\rm 1},
    Yibo Fan\textsuperscript{\rm 1}
}
\affiliations{
    \textsuperscript{\rm 1}Fudan University
    \textsuperscript{\rm 2}DAMO Academy, Alibaba Group 
    \textsuperscript{\rm 3}Hosei University

}

\usepackage{bibentry}
\nocopyright
\begin{document}

\maketitle

\input{sec/0.Abstract}

%

\input{sec/1.Introduction}

\input{sec/2.Related_work}
\input{sec/3.Preliminaries}
\input{sec/4.Method}
\input{sec/5.Experiment}

\input{sec/6.Conclusion}

\bibliography{aaai25}
\end{document}

%% file: sec/0.Abstract.tex
\begin{abstract}
    Prevalent lossy image compression schemes can be divided into: 1) explicit image compression (EIC), including traditional standards and neural end-to-end algorithms; 2) implicit image compression (IIC) based on implicit neural representations (INR). The former is encountering impasses of either leveling off bitrate reduction at a cost of tremendous complexity while the latter suffers from excessive smoothing quality as well as lengthy decoder models. In this paper, we propose an innovative paradigm, which we dub \textbf{Unicorn} (\textbf{U}nified \textbf{N}eural \textbf{I}mage \textbf{C}ompression with \textbf{O}ne \textbf{N}number \textbf{R}econstruction). By conceptualizing the images as index-image pairs and learning the inherent distribution of pairs in a subtle neural network model, Unicorn can reconstruct a visually pleasing image from a randomly generated noise with only one index number. The neural model serves as the unified decoder of images while the noises and indexes corresponds to explicit representations. As a proof of concept, we propose an effective and efficient prototype of Unicorn based on latent diffusion models with tailored model designs. Quantitive and qualitative experimental results demonstrate that our prototype achieves significant bitrates reduction compared with EIC and IIC algorithms. More impressively, benefitting from the unified decoder, our compression ratio escalates as the quantity of images increases.
    We envision that more advanced model designs will endow Unicorn with greater potential in image compression. We will release our codes in \url{https://github.com/uniqzheng/Unicorn-Laduree}.

    \end{abstract}

%% file: sec/1.Introduction.tex
\section{Introduction}
\label{sec: intro}
Lossy image compression aims to reduce images to smaller intermediate representations and subsequently restore them with minimal information degradation. This crucial aspect of modern digital imaging has emerged from the necessity for efficient and reliable image storage and transmission to conserve capacity and bandwidth. 
From the perspective of compressed image representation, compression frameworks can be categorized into two types, i.e., explicit and implicit.



\begin{figure}[t]
\centering
\includegraphics[width=1.0\linewidth]{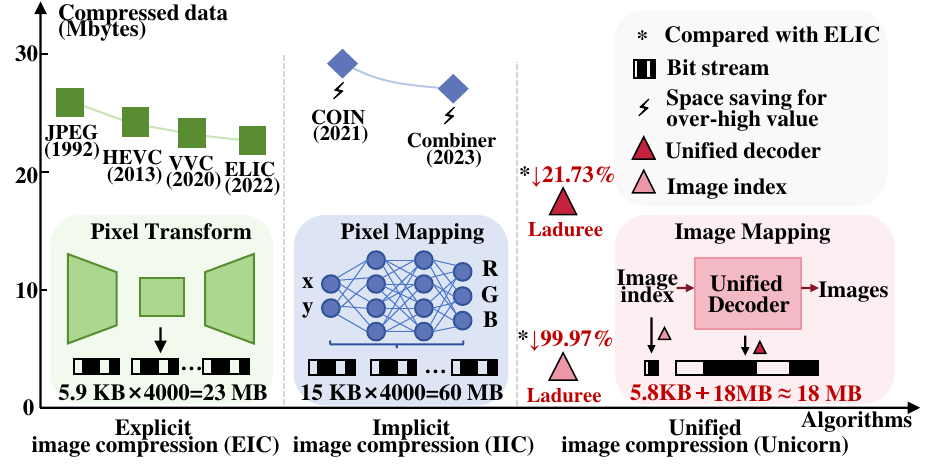}
\caption{Bitrates comparison among EIC, IIC, and Unicorn when compressing $4000$ images at the high perceptual quality ($\text{LPIPS} = 0.10$ for EIC and Uncorn while $0.35$ for IIC since it's hard to approach satisfactory perceptual quality.}
\label{fig:bitrate-model-comparison_toyexp}
\end{figure}

Previous explicit methods include traditional compression standards~\cite{wallace1991jpeg,sullivan2012overview,bross2021overview} and end-to-end neural compression algorithms~\cite{balle2018variational,mentzer2020high,he2022elic}.
They perform complex transformations and intricate redundancy removal in the pixel space, explicitly representing each image with a bitstream, known as Explicit Image Compression (EIC).
Limited to Lossy Minimum Description Length (LMDL) principle~\cite{madiman2004minimum,ma2007segmentation}, EIC algorithms are experiencing diminishing returns in bitrate reduction as model complexity significantly increases~\cite{bossen2021vvc,hu2021learning}, as illustrated in Figure~\ref{fig:bitrate-model-comparison_toyexp}. 
Continuing to progress in this direction, the model's size overhead might outweigh the benefits of bit stream reduction. 


To mitigate this, a recent line of work~\cite{dupont2021coin,strumpler2022implicit,guo2023compression}, so-called Implicit Neural Representation (INR)-based image compression (IIC), reformulates the problem of image compression as coordinate-pixel mapping. 
Specifically, it trains a unique neural network (NN) model to overfit the coordinate-to-pixel mapping of each image. 
Compression is achieved in two ways: first, 
IIC significantly reduces the explicit representation of each image from long bitstream to coordinates, which in this paper we regard as implicit representation, specified by height and width. 
Second, each image is represented with a dedicated compression model that extracts its spatial and structural redundancy into the neural network weights. The code lengths of height and width as well as the NN model constitute the final bitrates.
The simple implicit representation and tailored unique compression model remarkably reduce the model complexity~\cite{Ladune_2023_ICCV}. 

However, the coordinate-to-pixel mapping method focuses on learning a continuous function that efficiently represents image. This process may lead to losing fine details and textures, impacting overall perceptual quality. 
Moreover, the unique per image NN model ignores the similarity between images, resulting in inter-image information redundancy. 
As seen in Figure~\ref{fig:bitrate-model-comparison_toyexp}, IIC consumes much more bitrate than EIC to achieve relatively high perceptual quality.
To this end, we propose a brand-new image compression paradigm by obtaining compact representations both explicitly and implicitly.
We conceptualize the whole set of images to be compressed as index-image pairs and learn the inherent probability distribution among pairs in one lightweight NN model, thereby eliminating the inter-image information redundancy.
Additionally, in our design the decoding process is started from any random generated noise to guarantee the decoding process in pixel domain. Therefore, we can easily embrace the better perceptual quality benefit of EIC scheme. 
The index in our design represents the implicit representation, while any random generated noise corresponds to the explicit representation. 
Such one single NN model serves as the \textit{unified decoder} of a set of images, and the combined representations are called \textit{unified representation}. 
In that case, we reformulate the image compression problem as an image distribution sampling task which can be managed by one single index value. 
We name the new image compression paradigm as \textbf{U}nified \textbf{N}eural \textbf{I}mage \textbf{C}ompression with \textbf{O}ne \textbf{N}number \textbf{R}econstruction, dubbed \textbf{Unicorn}. Note that the code lengths of index numbers and the unified decoder model constitute the overall bitrates of the set of images.

Advancements in the NN model can fully exploit Unicorn's potential in rate-distortion performance. As a proof of concept, we propose a prototype by investigating a novel NN model based on the conditional latent diffusion model (LDM), which 
perfectly fits our requirements with randomly generated noise and conditional index.
We dub the prototype as \textbf{La}tent \textbf{d}iffusion-based \textbf{u}nified \textbf{re}pr\textbf{e}sentation (\textbf{Laduree}). As shown in Figure~\ref{fig:bitrate-model-comparison_toyexp}, Laduree outperforms prevalent EIC and IIC algorithms in overall bitrates saving, up to $21.73\%$ compared with ELIC~\cite{he2022elic} at the high perceptual quality ($\text{LPIPS}=0.10$). Notably, $99.97\%$ bitrate reduction can be achieved by transmitting one index number for one image reconstruction when the unified decoder is shared with the receiver.
We envision that more effective and efficient model designs in future work can further extend the compression potential of Unicorn.
Our contribution can be summarized as follows:
\begin{enumerate}
    \item We propose a novel image compression paradigm by conceptualizing images as index-image pairs and learning the inherent distribution of pairs in one NN model. 
    We dub the paradigm as Unicorn. 

    \item We propose a prototype of Unicorn based on advanced conditional LDM, which perfectly fits the requirements with built-in noise and conditional index. Subtle model designs tailored to Unicorn are comprehensively explored to obtain an effective and efficient unified decoder, which we dub Laduree.
    \item Quantitative and qualitative experiments demonstrate the superiority of Laduree on rate-distortion compression performance. More impressively, Laduree yields an increasing compression ratio as the number of compressed images increases, showcasing its great potential for large-scale image compression scenarios.  
\end{enumerate}

%% file: sec/2.Related_work.tex
\section{Related Work}
\subsection{Explicit image compression} In the last three decades, traditional image compression standards, such as JPEG~\cite{wallace1991jpeg}, JPEG2000~\cite{skodras2001jpeg}, HEVC intra~\cite{sullivan2012overview}, and VVC intra~\cite{bross2021overview}, have been refining a hybrid coding scheme based on handcrafted predictive coding and transform coding strategies, and explicitly represent images with bitstreams. These standard algorithms have been challenged by NN-based algorithms that non-linearly transform the pixels to the latent features and encode these features further to represent images explicitly. These algorithms jointly optimize the perceptual quality and bitrates in an end-to-end manner with Generative Adversarial Networks (GAN)~\cite{agustsson2019generative,mentzer2020high} and Variational Auto-Encoders (VAEs) models~\cite{balle2018variational,he2021checkerboard,he2022elic}. 

\subsection{Implicit image compression} Recent advancements have extended Implicit Neural Representations (INR) to image compression, where a dedicated NN model is overfitted in learning the mapping of coordinate-pixel value within an image. Therefore, one image is implicitly represented by coordinates, with which the dedicated NN model can output the corresponding image. After the pilot work of COIN~\cite{dupont2021coin} in this field, succeeding works enhance the rate-distortion performance by introducing meta-learned initialized model weights~\cite{strumpler2022implicit}, rate-distortion optimization~\cite{guo2023compression},  hierarchical latent representation~\cite{ladune2023cool}, and learning the entropy model~\cite{ladune2023cool}. 

\begin{figure*}[t]
\centering
\includegraphics[width=0.95\textwidth]{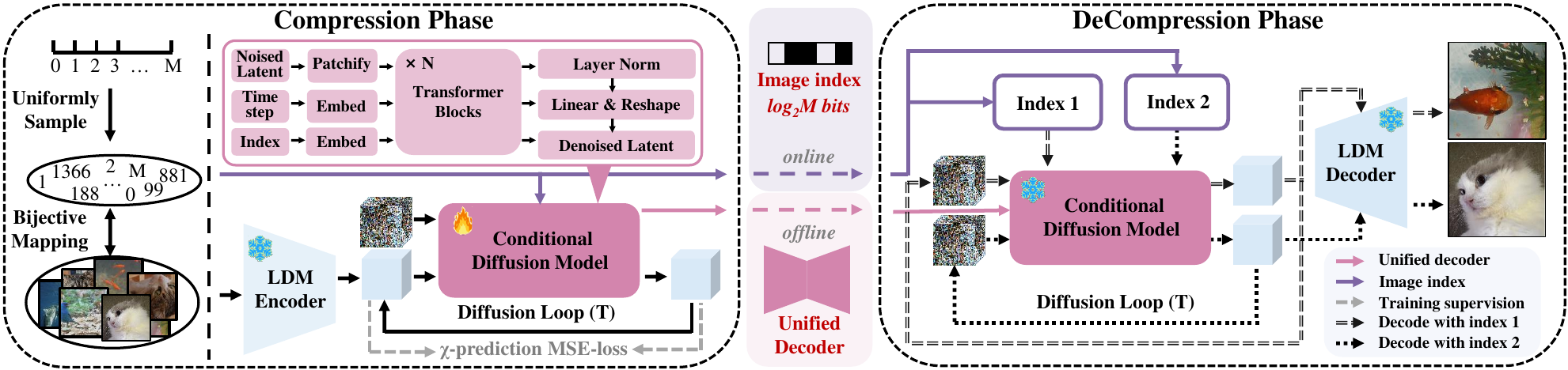}
\caption{Overall framework of the proposed paradigm Unicorn specified by the proposed prototype Laduree.}
\label{fig:proposed_framework}
\end{figure*}

\subsection{Latent Diffusion Models}
\label{rela:ldm}
Denosing Diffusion Probability Models (DDPM)~\cite{ho2020denoising}, referred to diffusion models hereafter, assume a parameter-free forward noising process where small portions of noise are applied to the input images. In the reverse process, diffusion models are trained to invert forward process corruptions. When the training is complete, one can initialize random noise and sample step by step to generate images. 
Latent Diffusion Models (LDMs)~\cite{rombach2022high} represent an advanced class of diffusion model variants, which transform images into a latent space via VAE before applying a diffusion process. Such a two-step process efficiently balances the perceptual quality of image generation with computational demands. Among them, Diffusion Transformers (DiT)~\cite{peebles2023scalable} is a pivotal advancement of LDMs, which enhances the scalability and efficiency of diffusion processes by integrating VAE and Transformer.

%% file: sec/3.Preliminaries.tex
\section{Preliminaries}
We formulate EIC and IIC schemes from the perspective of lossy data compression in information theory. 
\subsubsection{Background of Lossy Data Compression}
Consider the source data $X_1^n$ to be lossily compressed. Let $\rho(X_1^n, D)$ denote the reconstructed data with distortion $D$, and $P$ denote a probability distribution on the reconstructed data. $P$ is precisely correlated with the compression algorithm~\cite{kontoyiannis2002arbitrary}.
As it turns out in Lossy Minimum
Description Length (LMDL) principle~\cite{madiman2004minimum,ma2007segmentation}, given a family of probability distributions $\{P_\theta;\theta \in \Theta\}$, the optimal lossy compression occurs when an optimal probability distribution is found to compress the source data with as few bits as possible, including the cost of describing the distribution itself, which can be denoted as:
\begin{equation}
    \hat{\theta}^{\text{LMDL}} = \arg\min_{\theta \in \Theta} \left[ -\log_2 P_\theta \left( \rho(X_1^n, D) \right) + K(\theta) \right],
\end{equation}
where the first term indicates idealized lossy Shannon code lengths, and the second term measures the complexity of modeling probability distribution $P_\theta$~\cite{kolmogorov1998tables}. 

\subsubsection{Explicit image compression} In lossy image compression, we consider compressing a set of $M$ images $S := \{I_i\}_{i=1}^M$ with distortion $D$. A typical EIC algorithm $T$ consists of pixel transformation $\phi$ and variable entropy coding $\epsilon$, thus the description length (DL) of $S$ can be derived as:
\begin{equation}
\label{eq: mdl_precomp_oneimg}
    L^{EIC}_\text{DL} (S) = \sum_{i=1}^M -\log_2 P_\epsilon(\widetilde{V_i}) + K(\phi,\epsilon),
\end{equation}
wherein $\widetilde{V_i}$ indicates quantized variables obtained by performing transformation $\phi$ on pixels. The first term in Equation~\ref{eq: mdl_precomp_oneimg} calculates the code lengths of explicit representations for images while the second term measures the complexity of the general decoder. The detailed analysis of the EIC scheme is elaborated in Appendix 1.1. 

\subsubsection{Implicit image compression}
In IIC scheme, each image is compressed by overfitting a dedicated NN model $\psi$ for the mapping pixel coordinates $C$ to RGB intensities $E$. Therefore, the DL can be computed as:
\begin{equation}
\label{eq: mdl_inrcomp_allimg}
    L^{IIC}_\text{DL} (S) = \sum_{i=1}^M -\log_2 P_{\psi_i}(E_i|C_i) + \sum_{i=1}^M K(\psi_i).  
\end{equation}
In the first term, implicit coordinates $C$ can be derived from the width and height of images. The second term can be measured by the encoded weights of NN models~\cite{deletang2023language}. Hence, $2\times M$ numbers as well as $M$ dedicated NN models for each image are encoded to make up the overall bitrates. Information loss occurs in the approximation of loss function in model training. The detailed analysis of the IIC scheme is elaborated in Appendix 1.2.



%% file: sec/4.Method.tex
\section{Method}

%
\subsection{Proposed Paradigm}
We propose a novel image paradigm by conceptualizing images as index-image pairs and learning the inherent probability distribution with one NN model. Starting from random noise, the model only takes an extra index number as input to reconstruct the corresponding image with satisfactory perceptual quality. We name the paradigm as Unicorn.

\subsubsection{Formulation} 
Given that the overfitting problem of fake/random labels can be easily handled by neural networks~\cite{zhang2021understanding}, we initialize a set of fake/random index uniformly sampled in $\{1,...,M\}$ to construct a bijection function with the image set, denoted as $\widetilde{S} := \{(I_i, Y_i)\}_{i=1}^M$. 
A NN model $q$ which can generate images from a randomly generated noise, is leveraged to learn the bijection mapping following conditional distribution $P(I_i|Y_i)$, manifesting as the unified decoder. The probability distribution is identically equal to a uniform distribution $P(I|Y)=\frac{1}{M}$~\cite{blier2018description}.
Therefore, the DL of compressing $\widetilde{S}$ can be computed as:
\begin{equation}
\label{eq: dl_ourscheme}
  \begin{aligned}
    L^{Unicorn}_\text{DL} (\widetilde{S}) &= -\log_2 P_q(I|Y) + K(q)  \\
      &= M \log_2 M + K(q). \\
  \end{aligned}
\end{equation} 

\begin{figure*}[t]
\centering
\includegraphics[width=0.85\textwidth]{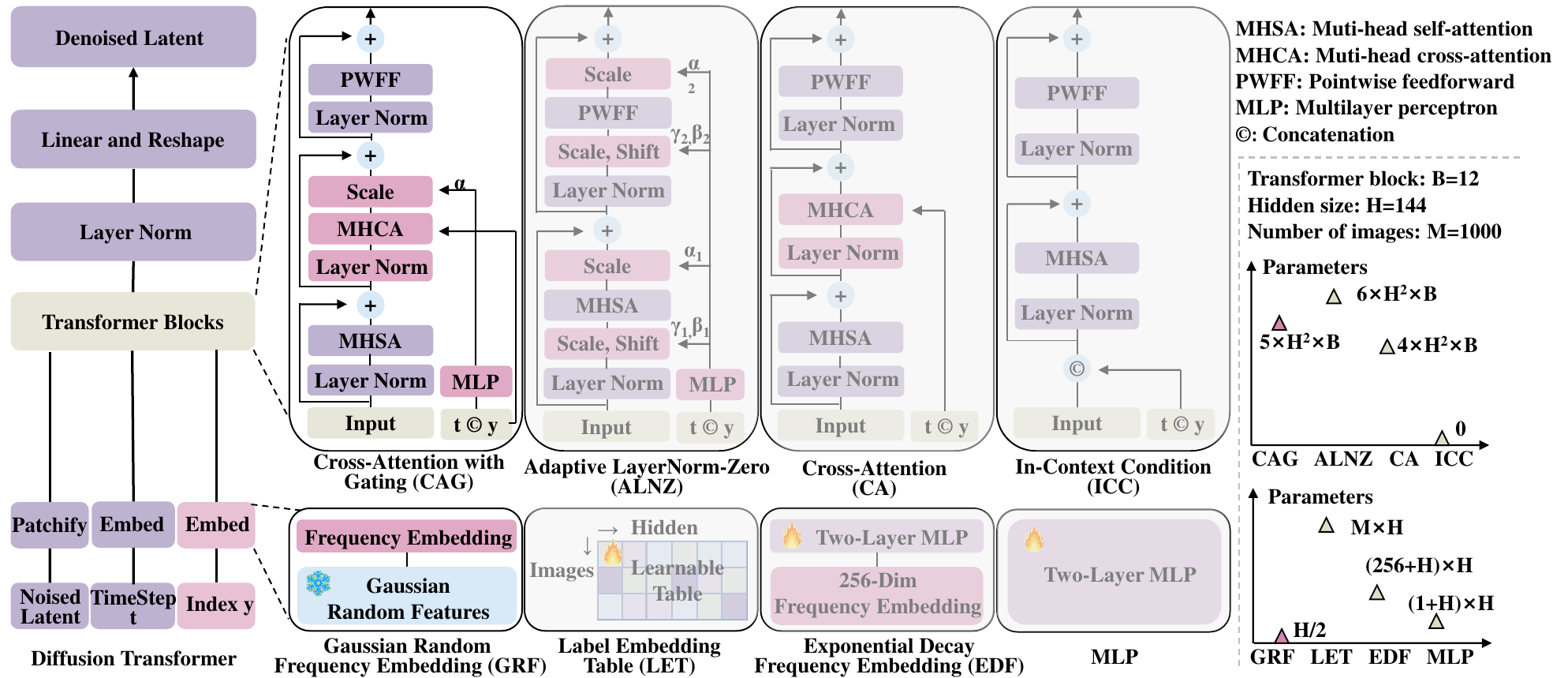}
\caption{Design space explorations on various manners for index embedding and condition within Transformer-based denoising model, with parameters introduced by each manner compared in the right bottom.}
\label{fig:diffusion_space_design}
\end{figure*}

\subsubsection{Comparison with EIC and IIC scheme.} 
Compared with EIC, tremendous bitrate reduction can be achieved in Unicorn by transmitting one index number for one image reconstruction when the unified decoder is shared with the receiver. Given that the code length of indexes is dependent on the number of images to be compressed, we quantitatively conduct bitrates comparison between Unicorn and ELIC. It turns out that the low bitrate transmission superiority in Unicorn can be maintained when compressing images of normal magnitude (see details in Appendix 1.3).
Compared with IIC, Unicorn consumes only one index number to reconstruct one image from the unified decoder $q$, while two for the IIC scheme. Moreover, the unified decoder $K(q)$ of Unicorn can achieve lower bitrates by eliminating the inter-image redundancy, compared with independent mapping learning for each image in IIC. 
\subsection{Proposed Prototype}
The core design lies in the NN model with two critical design requirements: 1) generate high-quality images from explicit random noise with index as extra input and 2) A small model size that allows for low bitrates. As a proof of concept, we investigate a subtle prototype based on latent diffusion models as depicted in Figure~\ref{fig:proposed_framework}. We name it as \textbf{Laduree}. 

\subsubsection{Overall implementation.} 
With LDM introduced in Section~\ref{rela:ldm} and preliminaries provided in Appendix 2.1, we focus on the overall implementation of Laduree.

As shown in Figure~\ref{fig:proposed_framework}, in the image compression phase, we uniformly sample index set Y in $\{1,...,M\}$ and bijectively map the index set to the image set with $M$ images. Secondly, following DiT~\cite{peebles2023scalable}, we use the off-the-shelf pre-trained VAE encoder to generate the latent features $Z$ from the images and construct the dataset $S := \{(Z_i, Y_i)\}_{i=1}^M$ following the uniform distribution $P(Z|Y)$. Then we train the conditional latent diffusion model on the dataset $S$, where $Z$ is generated conditioned on $Y$. After training, model compression strategies can be introduced to further reduce the model size. In the image decompression phase, the index controls the latent denoising diffusion process, which are then input to the VAE decoder to generate the corresponding image. Note that the pre-trained VAEs can generalize to any latent diffusion models, thus the bitrate consumption in compressing images only comes from the encoded weights of the latent diffusion model. The overall bitrate cost can be controlled by varying the number of parameters of model weights and adjusting the ratio of model compression. Notably, considering once the latent diffusion model is offline shared with the receiver, the online transmission cost is only $\log_2 M$ bits per image.

\subsubsection{Efficiency through similarity}
Laduree enjoys significant efficiency gains when compressing images with similar semantics. 
In the training stage, the ground-truth distribution learned by the conditional diffusion models in the reverse process is denoted by $q^*(x_{t-1}|x_t, x_0, Y)$. 
Due to the bijective mapping between the index $Y$ and the latent features $x_0$, this distribution effectively reduces to $q^(x_{t-1}|x_t, x_0)$, akin to an unconditional latent diffusion model. The proof is provided in Appendix 2.2.
As the diffusion model learns not only the bijective mapping but also the underlying distribution within latent features, it becomes particularly advantageous for the model to operate more efficiently when images exhibit high similarity.
We deem it as an advantage for compressing similar images with a small diffusion model.

\subsubsection{Index condition and embedding}
Different from the typical conditional generation task where diffusion models are trained to generate a set of images from the same category conditioned on one class label, our model is trained to overfit the bijectively paired image-index data. Driven by the trade-off on rate-distortion performance, we comprehensively conduct non-trivial explorations on index condition and embedding tailored for our paradigm to yield light yet effective diffusion model, as shown in Figure~\ref{fig:diffusion_space_design}.

For index condition, we propose the cross-attention with gating block (CAG) by adding additional dimension-wise scaling parameters $\alpha$ to the residual connections after the cross-attention block, with only introducing a few parameters based on vanilla Multi-head cross-attention (CA)~\cite{vaswani2017attention}. Note that In-context condition (ICC), Multi-head cross-attention (CA), and Adaptive layer norm with zero initialization (ALNZ) have been previously evaluated to gain increasing performance in image generation~\cite{peebles2023scalable}, yet also introducing increasing parameters. Our proposed CAG can achieve competitive performance with ALNZ with less model parameters.

With the effective conditioning manner, we resort to a simple yet effective embedding method for the index, which is Gaussian random Frequency embedding (GRF) with a few training-free parameters. Ever used in early diffusion models~\cite{rombach2022high}, GRF's parameters increase only linearly with hidden sizes. The widely-used robust embedding manner for variable timesteps in recent diffusion models~\cite{peebles2023scalable} is the exponential decay frequency embedding~\cite{vaswani2017attention} (EDF). However, its parameters increase quadratically with hidden sizes. Additionally, the label embedding table (LET) in typical conditional image generation model~\cite{peebles2023scalable} introduces trainable parameters increasing linearly with the number of images, which is unacceptable when compressing large-scale images in our paradigm. Multilayer perceptron (MLP) is included simply as a reference in comparing parameters. The GRF leveraged in Laduree yields competitive performance with EDF with much less parameters.

\subsubsection{Normalization for latent features.}
The latent features obtained by VAE follow a non-standard, zero-mean normal distribution. In typical LDM~\cite{peebles2023scalable}, the latent features are pre-processed to follow a standard normal distribution $N(0,1)$. We assume that when learning the index-image bijective mapping, a more concentrated and controlled latent space can reduce the variance in the generated outputs, making them more consistent with the conditioning index. Therefore, we scale the normal distribution to obtain a standard deviation of $\frac{1}{3}$, following $N(0,{\frac{1}{9}})$. This adjustment ensures that the majority of the data is concentrated within a more focused interval of $[-1, 1]$, according to $3\sigma$ rule. 

\subsubsection{Model compression}
Existing model compression strategies such as retraining, pruning, and quantization can be employed for lower bitrates. In Laduree, we adopt simple but efficient quantization to compress the model. Note that even without entropy coding or learning a distribution over model weights, Laduree can deliver promising rate-distrotion performance.
Concretely, all the learnable parameters of the trained network are quantized from 32-bit to (1+$e$+$m$)-bit in floating-point representation. We arrange the floating-point bits following the IEEE-754 standard~\cite{kahan1996ieee}. The sign bit is reserved, the exponent is clamped to the range $[-2^{e},2^{e}-1]$, and the mantissa is truncated to $m$ bits.

%% file: sec/5.Experiment.tex
\begin{figure*}[!t]
\centering
\includegraphics[width=0.9\textwidth]{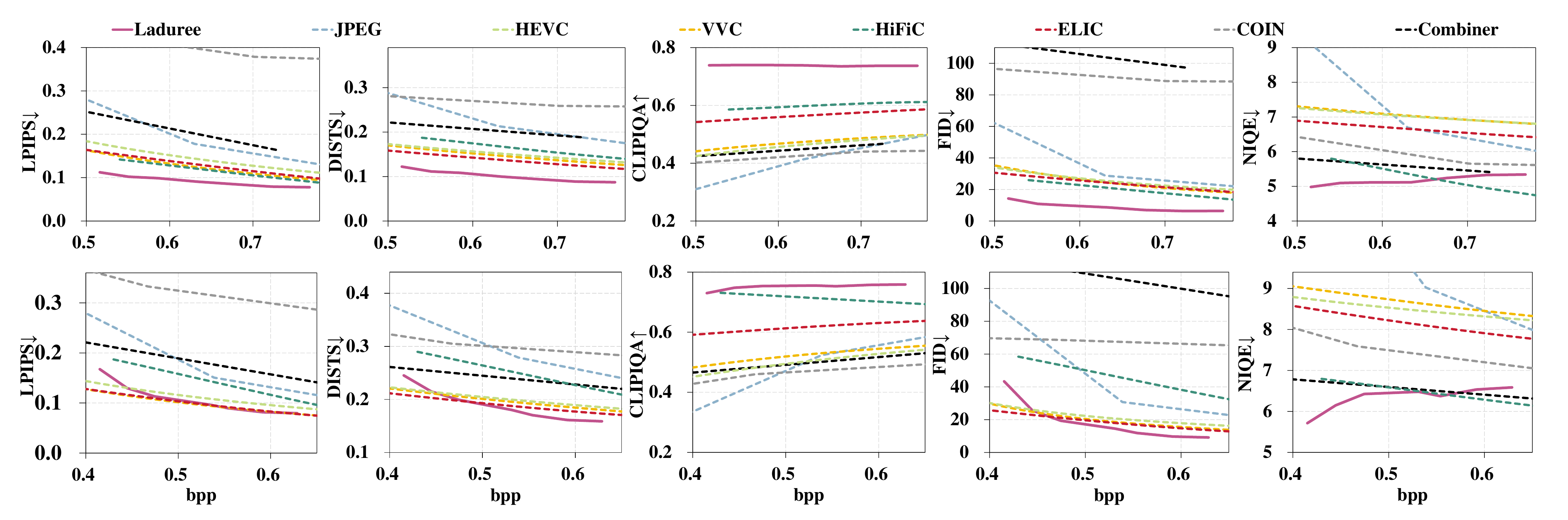}
\caption{RD curves of evaluated image compression models on CAT (top row) and HYBRID (bottom row) when compressing $4000$ images.}
\label{fig:cat_hybrid_rd-curve}
\end{figure*}

\begin{figure}[t]
\centering
\footnotesize
\includegraphics[width=0.9\linewidth]{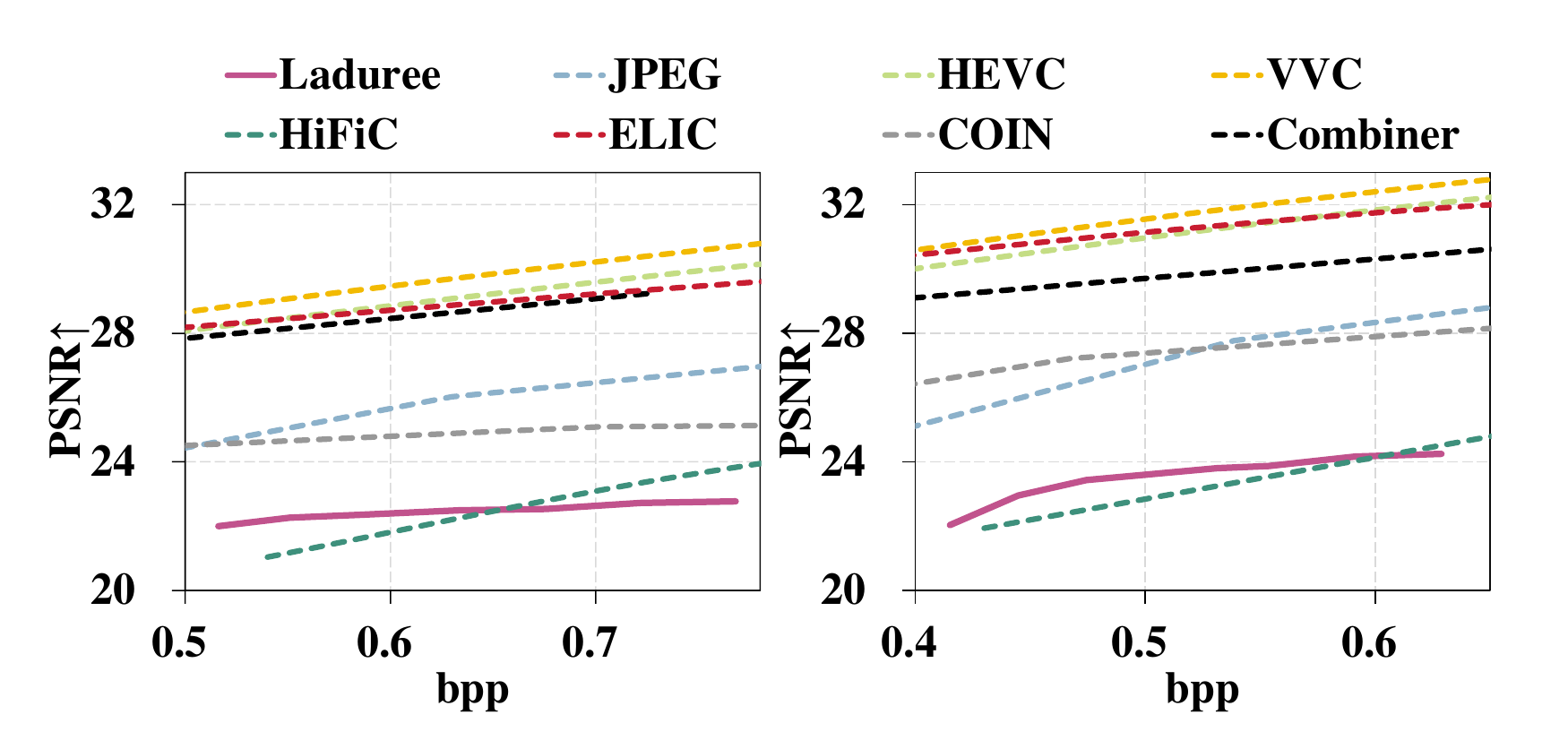} 
\caption{RD performance comparison in terms of PSNR.}
\label{fig:rd_curve_psnr}
\end{figure}

\begin{figure}[t]
\centering
\includegraphics[width=0.9\linewidth]{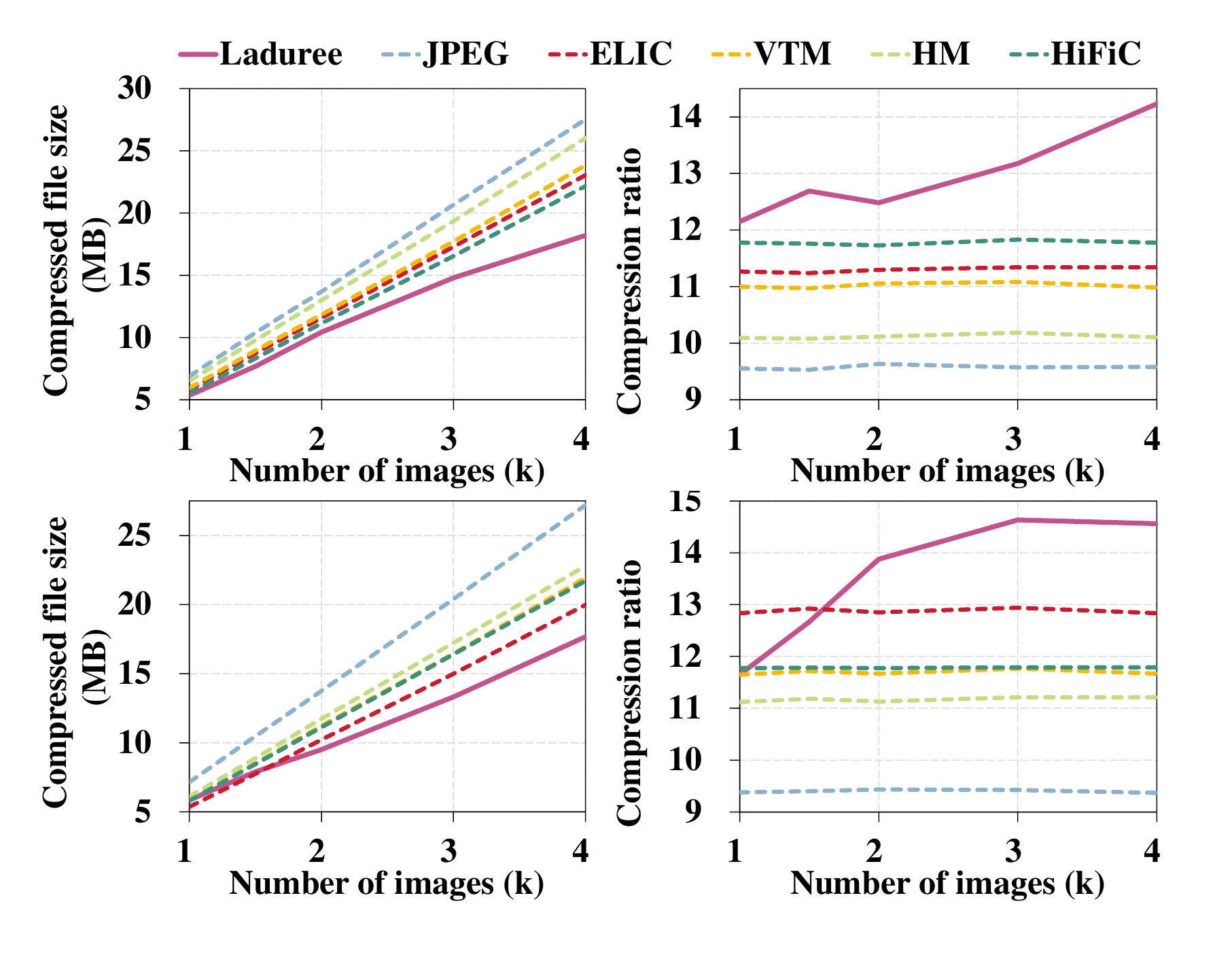}
\caption{Compressed file size and compression ratio of CAT (top row) and HYBRID (bottom row).}
\label{fig:storage_cat_hybrid}
\end{figure}

\section{Experiment}
\label{sec:exp}
\subsection{Experiment Setup}
\subsubsection{Dataset} 
We extract images from the training dataset of ImageNet~\cite{russakovsky2015imagenet} and downsample them to $256\times256$ resolution. 
We prepare a `CAT' image set containing $4000$ images from the various cat categories, and a `HYBRID' image set containing $4000$ images from 5 categories, 
each of 800 images, referring as $\text{CAT-}4000$ and $\text{HYBRID-}4000$ hereafter. Index numbers are uniformly sampled from $\{0,...,3999\}$, and bijectively mapped to the $4000$ images for both `CAT' and `HYBRID' image sets. We further divide each dataset into five subsets with the number of images in $\{1000,1500,2000,3000,4000\}$.
\subsubsection{Implementation details} 
To comprehensively evaluate the rate-distortion (RD) performance of our models under variable bitrates, we vary the number of weight parameters in the diffusion model and train $10$/$9$ diffusion models for the aforementioned subsets of CAT/HYBRID datasets, respectively. We fix the Transformer block depths $B=12$ and adjust the hidden sizes $H$ ranging from $108$ to $240$
as well as the quantization precision $W$ ranging from $32$ bits to $10$ bits.The model configurations of each subset are elaborately reported in Appendix 3.1. We name our model following $\text{Data-X}^{HX}_{WX}$ hereafter. For example, $\text{CAT-1500}^{H120}_{W14}$indicates a latent diffusion model compressing $1500$ images of CAT with Transformer hidden size $120$ quantized at $14$ bits. 

Following~\cite{nikankin2022sinfusion,yang2024lossy}, we train the latent diffusion model by predicting the clean latent features $x_0$ instead of noise $\epsilon$ with beneficial effects of satisfactory perceptual quality and less denoising timesteps. Compared to the $\epsilon$-model with hundreds of steps, Only $50$ timesteps are needed in Laduree. We use the Mean-Square-Error (MSE) as the loss function for predicting $x_0$. On NVIDIA A6000 GPUs, the model is trained for $50$ epochs optimized by Adam with the learning rate initialized as $2\times10^{-4}$ and halving every $10$ epoch.     

\subsubsection{Baseline models} For EIC, we include JPEG, HEVC Intra and VVC Intra as traditional baseline models.
HIFIC~\cite{mentzer2020high} and ELIC~\cite{he2022elic} are evaluated as neural baseline models, where the former is a GAN-based model towards high perceptual quality and the latter is a VAE-based model.
For IIC, we evaluate COIN~\cite{dupont2021coin} and Combiner~\cite{guo2023compression}. The configuration details of these baseline models are introduced in Appendix 3.2.

\subsubsection{Performance Evaluators} 
We adopt a comprehensive set of perceptual quality metrics that are highly consistent with human vision system. Following~\cite{mentzer2020high,careil2023towards}, we include LPIPS~\cite{zhang2018unreasonable} and DISTS~\cite{ding2020image} for perceptual similarity, FID~\cite{heusel2017gans} for realism, CLIPIQA~\cite{wang2023exploring} for aesthetics, and NIQE~\cite{6353522} and BRISQUE~\cite{mittal2012no} for naturalness. We also use PSNR to measure fidelity for compression. Detailed introductions about quality metrics can be seen in Appendix 3.3. For rate computation, bits-per-pixel (bpp) is used to measure the average bits per pixel required to represent images. 

\subsection{Rate-Distortion Performance}
\begin{figure*}[!t]
\centering
\includegraphics[width=0.85\textwidth]{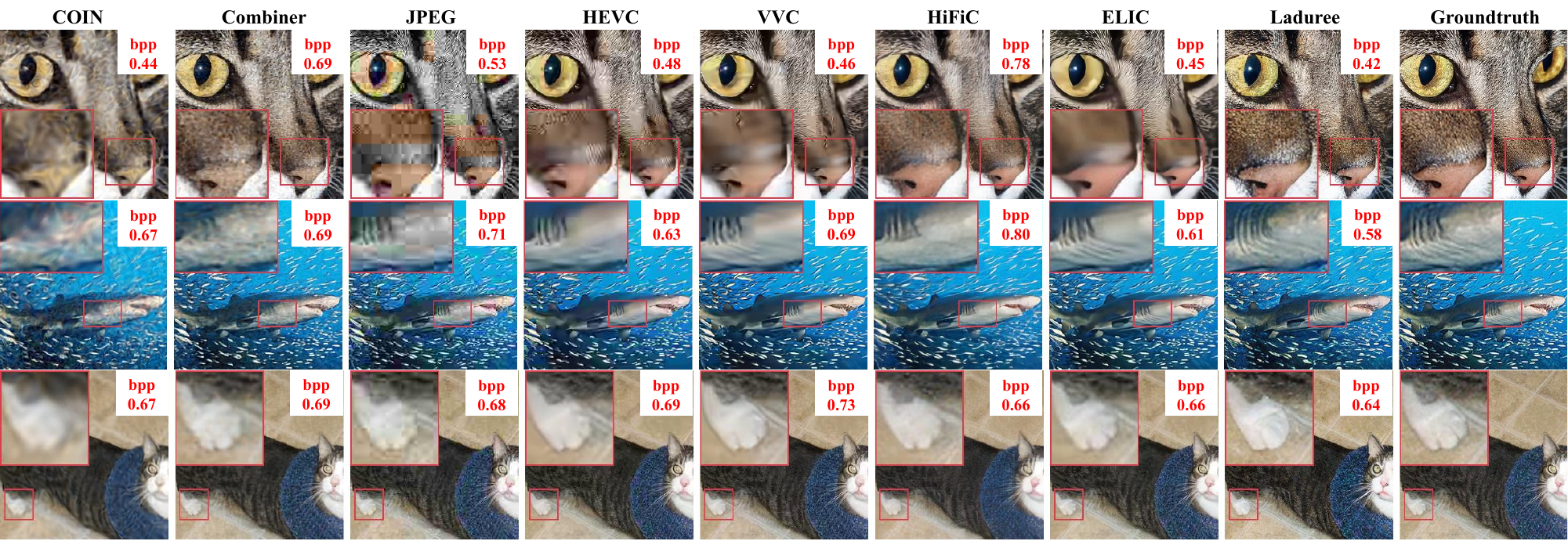}
\caption{Visual comparison among evaluated image compression models. Better zoom in.}
\label{fig:visual_comparsion}
\end{figure*}

Firstly, we provide the overall rate-distortion performance comparison among baseline models and our models fixed with GRF embedding and CAG conditioning.
Figure~\ref{fig:cat_hybrid_rd-curve} quantitatively illustrates the rate-distortion curves of evaluated compression algorithms on $\text{CAT-}4000$ (top row) and $\text{HYBRID-}4000$ (bottom row), wherein Laduree with configurations as $\text{CAT-4000}^{H208/224/240}_{W14/15/16}$ and $\text{HYBRID-4000}^{H192/204/216}_{W14/15/16}$ are evaluated. The RD performance comparison results when compressing images of 1000/1500/2000/3000 as well as BRISUQE results are reported in Appendix 4.1. From Figure~\ref{fig:cat_hybrid_rd-curve} it can be seen that Laduree significantly outperforms baseline models on delivering highly aesthetic and natural images with lower bitrates on both CAT and HYBRID datasets. Laduree shows remarkable superiority in high perceptual quality of similarity and realism on the CAT dataset while delivering competitive performance on HYBRID compared with the state-of-the-art. It shows that Laduree is capable of eliminating the inter-image similarity and yielding lower bitrates for images with similar semantics, showing its great potential in large-scale customized image compression applications.
Figure~\ref{fig:rd_curve_psnr} shows that Laduree can reach competitive PSNR values compared with HIFIC and maintain an acceptable fidelity when compressing images with high perceptual quality, which is consistency with the well-known rate-distortion-perception trade-off~\cite{blau2019rethinking}.
Figure~\ref{fig:visual_comparsion} shows the visualization comparison results. As can be seen, our models deliver more visually pleasing images with detailed textures at lower bitrates. See more visualization comparison in Appendix 4.2.

\subsection{Unique bitrate superiority}

Figure~\ref{fig:storage_cat_hybrid} shows the comparison of compressed file size and compression ratio at high perceptual quality on CAT ($\text{LPIPS} \leq 0.1$, top row) and HYBRID ($\text{LPIPS} \leq 0.08$, bottom row) image sets. 
On $\text{CAT-}2000$, Laduree achieves smaller compressed size, $9.90\%$ and $6.30\%$ over ELIC and HiFiC, respectively. When the quantity of images increases, Laduree can impressively achieve even more bitrates savings, manifesting as $21.73\%$ and $17.87\%$ savings than ELIC and HIFIC on $\text{CAT-}4000$, respectively. This unique bitrate superiority related to the quantity of images is consistent with the HYBRID dataset. 
Benefitting from eliminating the
inter-image redundancy in the unified decoder, Laduree significantly improves compression ratios and reduces file sizes as the number of images increases. In contrast, baseline models compress images independently, resulting in constant compression ratios and linear increases in file size. It demonstrates our unique superiority in large-scale image storage.


\begin{table}[t]
\centering
\fontsize{7pt}{9pt}\selectfont
\setlength{\tabcolsep}{3pt}
\caption{Performance comparison on latent pre-processing.}
\label{table:scaling_factor}
\begin{tabular}{c|ccccccc}
\toprule
\multirow{2}{*}{\begin{tabular}{c} Latent \\ Standard Deviation \end{tabular}} & \multicolumn{3}{c}{\centering CAT-1000} & &  \multicolumn{3}{c}{\centering HYBRID-1000}\\ 
 \cline{2-4}  \cline{6-8} \\ [-1.em]
& LPIPS & FID & NIQE && LPIPS & FID & NIQE\\
 \midrule
 1 & 0.16& 54.65& 12.65 &  & 0.15 & 29.21 & 13.01\\
 $1/3$ & \textbf{0.08} & \textbf{12.08}& \textbf{6.54}& & \textbf{0.06} & \textbf{14.24}& \textbf{6.08}\\
\bottomrule
\end{tabular}
\end{table}

\subsection{Design space exploration}
\label{exp:design_space_exploration}

\subsubsection{Normalization for latent features}
Table~\ref{table:scaling_factor} tabulates performance comparison between two normalization manners in latent pre-processing, where Laduree with configurations $\text{CAT-1000}^{H144}_{W16}$ and $\text{HYBRID-1000}^{H120}_{W16}$ are reported. As it can be seen, in our paradigm, the diffusion model can benefit from the less dispersed latent space and generate images with much better perceptual quality than the typical latent space with standard normal distribution.


\begin{figure}[t]
\centering
\footnotesize
\includegraphics[width=1\linewidth]{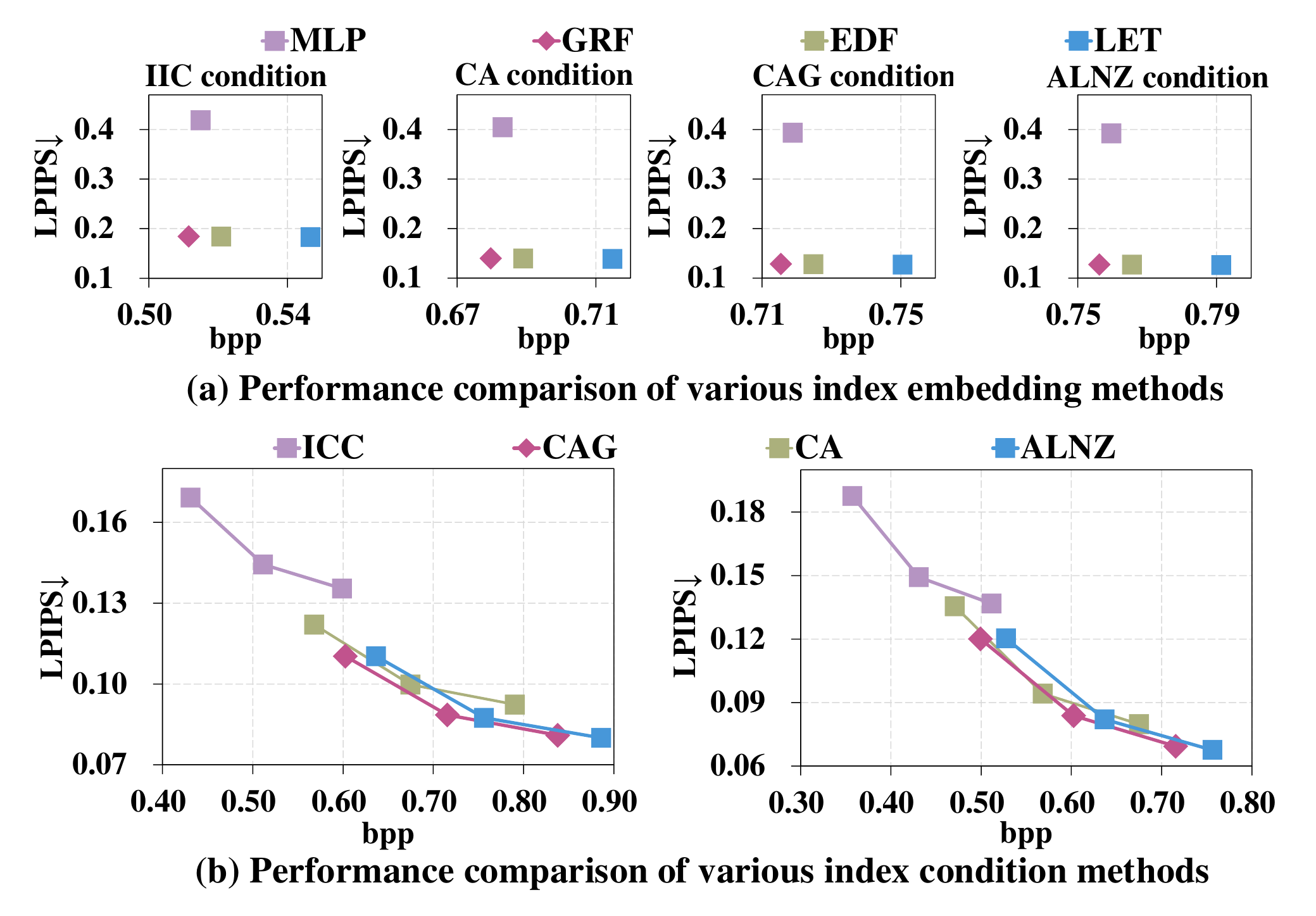} 
\caption{RD performance comparison of index embedding and condition.}
\label{fig:exp_embedding_condition}
\end{figure}

\begin{figure}[t]
\centering
\footnotesize
\includegraphics[width=0.9\linewidth]{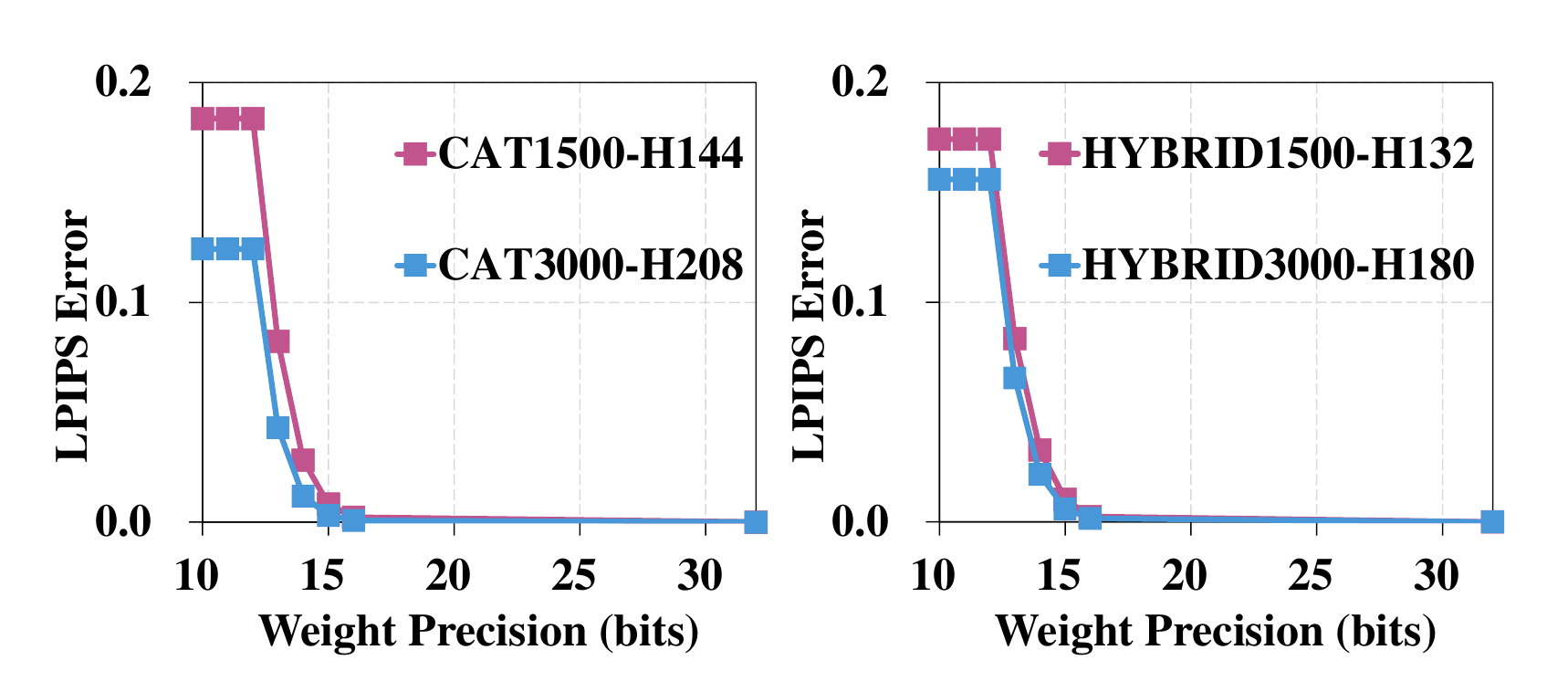} 
\caption{Robustness evaluation on weight quantization.}
\label{fig:exp_quantization}
\end{figure}

\subsubsection{Index embedding and condition}

We evaluated various embedding methods of index numbers in the latent diffusion model. Figure~\ref{fig:exp_embedding_condition} (a) showcases the compression performance of Laduree with configuration $\text{CAT-1500}^{H144}_{W16}$. Four embedding methods are evaluated on all four condition methods, including ICC, CA, proposed CAG, and LET. In every condition manner, GRF embedding outperforms the other embedding methods with the the least bitrates while yielding high perceptual quality.

We also evaluated various condition methods for label numbers with the embedding method fixed at the GRF embedder. Figure~\ref{fig:exp_embedding_condition} (b) shows the compression performance of our models with configurations $\text{CAT-1500}^{H132/144/156}_{W16}$ (left) and $\text{HYBRID-1500}^{H120/132/144}_{W16}$ 
 (right). It shows that the proposed CAG outperforms other condition methods, yielding less bitrates with better perceptual quality.



\subsubsection{Robustness on weight quantization}
Figure~\ref{fig:exp_quantization} depicts the quantization error of Laduree with configurations $\text{CAT-1500}^{H144}$ and $\text{CAT-3000}^{H208}$ shown in the left, and $\text{HYBRID-1500}^{H132}$ and $\text{HYBRID-3000}^{H180}$ shown in the right. Laduree maintains acceptable quality loss at $14\text{-bit}$ precision, achieving a rate of $2.3$ in model size compared with that in training precision of $32\text{-bit}$. Moreover, our framework shows increasing robustness to weight quantization as the model gets larger for more images, which again demonstrates the superiority of large-scale compression.


%% file: sec/6.Conclusion.tex
\section{Conclusion}

We proposed a new image compression paradigm by conceptualizing the whole image set as index-image pairs and learning the inherent conditional distribution in an NN model. We develop a prototype based on LDM with non-trivial design explorations. Comprehensive experiments demonstrate promising results of our framework against prevalent algorithms. We anticipate that more advanced model designs can endow the paradigm with greater potential.